\pdfoutput=1

\documentclass[11pt]{article}

\usepackage[preprint]{acl}
\usepackage{longtable}
\usepackage{geometry}
\usepackage{lscape}
\usepackage{times}
\usepackage{latexsym}
\usepackage{soul} 
\usepackage{xcolor}
\usepackage{tcolorbox}
\usepackage{multirow}

\usepackage[T1]{fontenc}

\usepackage[utf8]{inputenc}

\usepackage{microtype}

\usepackage{inconsolata}

\usepackage{graphicx}

\title{Current State-of-the-Art of Bias Detection and Mitigation in Machine Translation for African and European Languages: a Review}

\author{Catherine Ikae \\
  Applied Machine Intelligence \\
  Bern University of Applied Sciences \\
  Biel, Switzerland \\
  \texttt{catherine.ikae@bfh.ch} \\\And
  Mascha Kurpicz-Briki \\
  Applied Machine Intelligence \\
  Bern University of Applied Sciences \\
  Biel, Switzerland \\
  \texttt{mascha.kurpicz@bfh.ch} \\}

\begin{document}
\maketitle
\begin{abstract}
Studying bias detection and mitigation methods in natural language processing and the particular case of machine translation is highly relevant, as societal stereotypes might be reflected or reinforced by these systems. In this paper, we analyze the state-of-the-art with a particular focus on European and African languages. We show how the majority of the work in this field concentrates on few languages, and that there is potential for future research to cover also the less investigated languages to contribute to more diversity in the research field.
\end{abstract}

\section{Introduction}
Bias in Natural Language Processing (NLP) technologies has over the last years become an important subject of research. It was been shown that societal stereotypes are included in NLP models (e.g., \citet{bolukbasi2016man} \citet{caliskan2017semantics} \citet{wilson2024gender}). The bias can happen at different stages of the NLP pipeline \citet{hovy2021five}. Additionally, different forms of bias and in particular also intersectional bias can be present, making it challenging to detect and mitigate unwanted behavior caused by this in NLP models or applications. Furthermore, it has been shown that there is a disparity of language resources available for different languages (e.g., \citet{joshi2020state}). One particular case where bias can appear is machine translation (MT). For example, when translating the non-gendered sentence \emph{I am a professor.} from English to a language like German. In German, this sentence could translate to \emph{Ich bin ein Professor.} or \emph{Ich bin eine Professorin.}, depending on the gender of the speaker. The machine translation application will typically select one of them, and this choice might be impacted by biases in the underlying model. A case study has shown that in Google Translate, even if not expecting a 50:50 gender distribution, the machine translation engine yielded male defaults much more frequently than it would be expected from the corresponding demographic data \cite{prates2020assessing}.

Whereas different work has recently discussed the matter of bias in machine translation from different perspectives (e.g., \citet{savoldi2021gender}, \citet{gallegos2024bias}), there is a lack of overview on the languages involved in the current state-of-the-art research in this field. In this paper, we give an overview over the state-of-the-art of bias detection and mitigation in machine translation, with a focus on research involving European and African languages.

\section{Methods}
We have searched the literature in two steps. We first queried Web of Science using the query \emph{machine translation AND bias}, leading to 402 results. Those were downloaded as an excel file for further processing. We then used a script to filter title and abstract, to identify papers where at least one language from a list of African and European languages (full list in Appendix \ref{sec:appendix1}) appears in it. The list of languages was generated using ChatGPT, in a two step approach (see Appendix \ref{sec:appendix1} for details). After this filtering, we had 11 papers for African languages and 39 papers for European languages. Those were then further analyzed by manual abstract screening by one of the authors. In unclear cases, the full-text of the paper was consulted and the cases were discussed in the team. 

While doing this analysis, the authors realized that relevant papers from the ACL Anthology were not included, even though ACL was selected as a source. We therefore added another search directly on the ACL Anthology. The queries were composed to use already the filter for the languages. The query is described in Appendix \ref{sec:appendix1}. The first 5 pages of results sorted by relevance were manually assessed. Duplicates were removed. 18 papers were considered relevant for the purpose of our study. 
The excluded and included papers and exclusion/inclusion reasons are listed in Appendix \ref{sec:appendix2}. 

After this search and selection process, we identified 9 relevant papers for African languages, and 25 papers for European languages, which will be discussed and compared in the next section.

\section{Results}
In reviewing the existing research on bias in machine translation (MT) found in our study, it becomes evident that the majority of studies focuses on a selection of few high-resource languages, often from Western Europe. Languages such as English, German, French, and Spanish dominate the research landscape, appearing consistently across numerous studies. From the languages in the query, German and Spanish are the most frequently studied, appearing in 14 papers and French in 7 papers. Italian, Hebrew, Arabic and Chinese also receives notable attention, appearing in 5 papers (an overview of languages is shown in Table \ref{tab:mt_bias_languages}.

The other language from the query that appeared 4 times in the reviewed papers include Russian, while those that appeared twice are Polish, Portuguese, Hungarian, Finnish, Estonian, Catalan, Czech, Korean, Hindi, Japanese and Hebrew. Polish and Russian are predominantly featured in studies addressing gender bias in machine translation, while Hungarian and Finnish are often discussed in the context of underrepresented European languages, highlighting their unique linguistic challenges. Yoruba and Hebrew are often linked to specific translation or gender-related biases.

Several languages from the query appear only once in the reviewed literature, indicating a limited focus on them. These include Mandarin, Turkish, Ukrainian, Bengali, Punjabi, Gujarati, Tamil, Icelandic, Marathi, Latvian, Romanian, Yoruba, Indonesian, Mongolian, and one unspecified language. Arabic and Swahili, despite their significance in non-European contexts, receive limited attention, with Swahili being particularly underrepresented in the research. Icelandic, though a unique case of a lesser-studied European language, also appeared in only one paper.

From the original query, many other languages, including Amharic, Tigrinya, Kabyle, Somali, and Hausa, are entirely absent from the reviewed research.

Some languages that were not part of the original query did appear in the reviewed studies. For instance, Mongolian, Malay, and Japanese were mentioned in a few papers, indicating that there is some effort to explore bias in less commonly studied languages.

\begin{table*}[h!]
\centering
\begin{tabular}{|p{0.5cm}|p{5cm}|p{8cm}|}
\hline
\# & Author(s) & Languages \\ \hline
1  & \citet{solmundsdottirmean2022} & Icelandic \\ \hline
2  & \citet{savoldi_gender_2021} & English, French, German, Spanish \\ \hline
3  & \citet{leetargetagnostic2023} & English, French, Spanish \\ \hline
4  & \citet{costa-jussa_fine-tuning_2020} & English, Spanish \\ \hline
5  & \citet{triboulet_evaluating_2023} & English, French, Italian, Turkish \\ \hline
6  & \citet{kocmi_gender_2020} & Czech, German, Polish, Russian \\ \hline
7  & \citet{vanmassenhove_machine_2021} & English, French, Spanish \\ \hline
8  & \citet{stewart_whose_2024} & French, Italian, Spanish \\ \hline
9  & \citet{attanasio_tale_2023} & English, German, Spanish \\ \hline
10 & \citet{levy_collecting_2021} & English, Arabic, Czech, German, Spanish , Hebrew, Italian, Russian, Ukrainian \\ \hline
11 & \citet{cabrera_gender_2023} & English, Hebrew \\ \hline
12 & \citet{mash_unmasking_2024} & English, Catalan \\ \hline
13 & \citet{vashishtha_evaluating_2023} & Hindi, Bengali, Marathi, Punjabi, Gujarati, Tamil, German, Japanese, Arabic, Spanish, Mandarin, Portuguese, Russian, Indonesian \\ \hline
14 & \citet{habash_automatic_2019} & Arabic \\ \hline
15 & \citet{stafanovics_mitigating_2020} & English, Latvian \\ \hline
16 & \citet{saunders_reducing_2020} & English, Spanish \\ \hline
17 & \citet{prates_assessing_2019} & English, Hungarian, Chinese, Yoruba \\ \hline
18 & \citet{ji_simple_2020} & Korean, Chinese, Mongolian \\ \hline
19 & \citet{elaraby_gender_2018} & English, Arabic \\ \hline
20 & \citet{moryossef_filling_2019} & English, Hebrew \\ \hline
21 & \citet{wisniewski_biais_2022} & German, English \\ \hline
22 & \citet{escude_font_equalizing_2019} & English, Spanish \\ \hline
23 & \citet{cho_towards_2021} & English, German, French \\ \hline
24 & \citet{kruger_outline_2022} & English, German \\ \hline
25 & \citet{campolungo_dibimt_2022} & English, Chinese, Russian, Italian, Spanish, German \\ \hline
26 & \citet{freitag_ape_2019} & English, French, Romanian, German \\ \hline
27 & \citet{costa-jussa_gebiotoolkit_2020} & English, Spanish, Catalan \\ \hline
28 & \citet{wang_survey_2023} & Chinese, Hungarian, Hindi, English, Japanese, Portuguese, Spanish, Arabic, Korean, Hebrew, Italian \\ \hline
29 & \citet{shi_approximating_2023} & English, German \\ \hline
30 & \citet{pulaczewska_cross-gender_2024} & English, Polish \\ \hline
31 & \citet{lu_challenge_2020} & English, German, Chinese \\ \hline
32 & \citet{liu_alleviating_2023} & English, German \\ \hline
33 & \citet{kaeser-chen_positionality-aware_2020} & Unspecified \\ \hline
34 & \citet{iluz_exploring_2023} & English, German, Spanish, Hebrew \\ \hline
\end{tabular}
\caption{Papers and Languages Studied in Machine Translation Bias.}
\label{tab:mt_bias_languages}
\end{table*}

We analyzed and compared the methodologies employed across the studies focused on bias detection and mitigation in machine translation, highlighting key approaches such as corpus-based evaluations, gendered translation benchmarks, and algorithmic bias quantification to assess how each method captures and mitigates biases across different languages and systems. This led to the following groups of papers.

\paragraph{Papers that focus on bias detection using pre-defined test suites, corpora, or experiments to measure gender bias in machine translations.} Bias detection often focuses on how systems handle gendered terms, such as pronouns and professions, or whether stereotypes are propagated in translations. For instance, \citet{solmundsdottirmean2022} focuses on Icelandic, where rich gender inflections in adjectives and nouns make it ideal for studying bias through inflectional errors. In contrast, \citet{prates_assessing_2019} evaluates English, which uses gendered pronouns, Hungarian, a gender-neutral language, Chinese, which has gender-ambiguous pronouns, and Yoruba, where noun classes can affect gender interpretations. Papers like \citet{kocmi_gender_2020} analyze Czech, German, Polish, and Russian—languages with strong gendered noun and pronoun systems, where incorrect gender agreement can clearly highlight bias. In Polish, for example, grammatical gender is deeply embedded in professions, making it a focus of \citet{pulaczewska_cross-gender_2024}. French, Italian, and Spanish are explored by \citet{stewart_whose_2024} to assess bias in the context of same-gender relationships, as these Romance languages require consistent gender agreement. Meanwhile, English and Catalan in \citet{mash_unmasking_2024} reveal biases related to gender omission. Lastly, \citet{costa-jussa_gebiotoolkit_2020} uses a multilingual approach, examining diverse languages to reveal how bias manifests across different grammatical systems.

\paragraph{Bias mitigation in machine translation, introducing innovative methods to reduce gender bias in translations.} These approaches generally focus on two main strategies: adjusting the training data and modifying the model itself. For example, dataset adjustments, such as balancing gender representation in English and Latvian, helped to address the rich gender agreement system in Latvian, which requires attention to grammatical gender inflections \cite{stafanovics_mitigating_2020}. Contrastive learning, applied in French, Spanish, and English, is effective for languages with consistent grammatical gender rules, particularly in Spanish, where gendered nouns demand balanced training \cite{leetargetagnostic2023}. Fine-tuning on gender-balanced datasets has also proven successful for languages like Spanish \cite{costa-jussa_fine-tuning_2020}. In multilingual settings, debiasing word embeddings, as seen in English and German, helps manage German's three-gender system, which requires a more precise approach than binary-gendered languages \cite{escude_font_equalizing_2019}. Furthermore, context injection, used in Hebrew and English, adds external gender and number clues to clarify ambiguities in pronouns, particularly in Hebrew, where gender-neutral pronouns are absent \cite{moryossef_filling_2019}. These methods illustrate how bias mitigation strategies are tailored to the specific linguistic properties of each language.

\paragraph{Addressing specific gender-related challenges such as pronoun translation, coreference resolution, and the handling of relationships in translations.} These works often introduce context-sensitive approaches or language-specific methodologies to improve gender accuracy. For instance, \citet{attanasio_tale_2023} and \citet{kocmi_gender_2020} focus on German, Spanish, and Czech, which have clear grammatical gender systems, using interpretability methods and evaluation metrics like WinoMT to enhance gender pronoun translation and coreference resolution. In Arabic, \citet{habash_automatic_2019} emphasizes the reinflection of verbs and nouns, utilizing neural models tailored to the complexities of Arabic grammar, where gender agreement is required across pronouns, verbs, and adjectives. Additionally, \citet{stewart_whose_2024} examines the translation of same-gender relationships in French, Italian, and Spanish, addressing the challenges of aligning gendered terms with social contexts in Romance languages. Finally, \citet{iluz_exploring_2023} investigates how the distribution of gendered terms in Hebrew and German training data affects translation accuracy, pointing out that tokenization can either exacerbate or reduce bias, depending on how gendered terms are processed during translation.

\paragraph{Efforts to tackle exposure bias in neural machine translation (NMT)}. This involves improving translation quality through methods, that enhance translation quality by simulating real-world inference conditions or augmenting data. The goal is to make models more robust to noise and difficult translations, which can indirectly influence bias detection and mitigation. For example, \citet{lu_challenge_2020} introduces dynamic sampling techniques, which expose models to challenging or erroneous inputs during training. In languages like Arabic and German, where morphology and grammatical gender are key factors, this approach may help address the noise created by gender inflections or verb agreements. Similarly, \citet{liu_alleviating_2023} uses contextual augmentation and self-distillation to enhance sequence prediction, which is particularly useful in languages with flexible word order, such as Czech or Russian, where sentence structure variations can introduce noise. Although not specifically targeting gender bias, \citet{shi_approximating_2023} applies causal models to improve translation quality evaluation, which can indirectly mitigate bias by reducing translation errors in languages like Spanish and French, where accurate gender agreement is critical. These techniques show how exposure bias mitigation can vary in its effectiveness depending on the linguistic properties of each language.

\paragraph{Data collection plays a critical role in bias mitigation, with some researchers developing gender-balanced corpora for training and evaluation, aiming to improve datasets for detecting and mitigating gender bias in machine translation.} The languages included in these corpora significantly affect the outcomes of bias detection. For instance, \citet{costa-jussa_gebiotoolkit_2020} focuses on constructing a gender-balanced corpus from Wikipedia biographies, covering English, French, Spanish, and other multilingual applications. These languages, which feature strong gender marking in nouns, pronouns, and adjectives, require careful data balancing to ensure equal representation of male and female forms. Similarly, \citet{levy_collecting_2021} emphasizes the importance of gender-balanced datasets in both English coreference resolution and general machine translation tasks. In languages like English, where gender is primarily expressed through pronouns, ensuring a fair representation of gendered terms helps improve translation fairness. Additionally, when applied to more gendered languages like French and Spanish, these datasets play a crucial role in detecting bias in how gender agreements are handled in translations. 

\paragraph{Various evaluation benchmarks and methods have also been developed to detect bias in machine translation.} These benchmarks address specific linguistic challenges such as word-sense disambiguation (WSD), pronoun translation, and the impact of automatic post-editing (APE) in amplifying biases. For example, \citet{campolungo_dibimt_2022} evaluates semantic bias in WSD across languages like English and Chinese, where word ambiguity is common, highlighting how biases can emerge from ambiguous word choices. WSD is particularly challenging in Chinese, where many words lack clear gender markers, making it easier for translation systems to misinterpret gender. In Japanese and Chinese, \citet{wang_survey_2023} explores gender biases where pronouns are frequently dropped, requiring translators to infer gender from context, which can often lead to biased outputs when gender-neutral terms are incorrectly assigned. This issue of dropped pronouns makes gender bias more likely to occur, especially in languages where the context must be carefully reconstructed to retain accuracy. Additionally, \citet{freitag_ape_2019} investigates how APE affects French, Romanian, and German translations, where gender agreement plays a significant role. APE can introduce or exacerbate biases in these languages by amplifying small errors in gender agreement, which can lead to skewed or biased translations. These evaluations reveal how linguistic features, such as pronoun usage and gender agreement, play a critical role in bias detection and mitigation across different languages.

\paragraph{Pedagogical frameworks.} To raise awareness of bias in machine translation, some researchers propose pedagogical frameworks aimed at equipping students and researchers with the tools to understand and address these issues across various languages. For example, \cite{kruger_outline_2022} combines data literacy with machine translation literacy, teaching students how to detect and mitigate bias in translations of languages such as French, German, and Spanish, where grammatical gender and gender agreement present unique challenges. This framework also helps students navigate gender-neutral languages, like Finnish or Chinese, teaching them how to handle pronoun ambiguity and cultural context. Meanwhile, \citet{kaeser-chen_positionality-aware_2020} emphasizes the importance of understanding how developers’ backgrounds and perspectives might introduce bias into machine translation systems, particularly in gendered translations and cultural contexts. 

\section{Discussion}

\subsection{Limitations of the State-of-the-Art}

A general limitation we encountered in our study was the limited or non-available research for many of the African and European languages from our queries. This generally shows that there is room for future research efforts for bias in machine translation detection and mitigation, taking additional languages into consideration.

The authors of the reviewed papers acknowledge several challenges that limit the scope and applicability of their findings. Common themes include the need to expand research beyond specific linguistic features, incorporating more inclusive gender constructs, improving scalability, refining evaluation metrics, and empirically testing theoretical frameworks.

A key limitation is that many studies focus on specific linguistic features or bias types, which can hinder the generalizability of their results across different languages or grammatical categories. For example, the work by \citet{solmundsdottirmean2022} focuses exclusively on adjectives and their gendered inflections in Icelandic, potentially overlooking other forms of bias related to verbs or nouns. Similarly, \citet{attanasio_tale_2023} focuses exclusively on pronoun translation and gender accuracy, which overlooks broader aspects of gender bias that might emerge in other word categories like gendered nouns. Other papers, like \citet{stewart_whose_2024}, focus specifically on same-gender relationships, limiting their conclusions to relational contexts and potentially missing biases in non-relational settings. Papers such as \citet{habash_automatic_2019} and \citet{habash_automatic_2019} also focus on specific grammatical phenomena (e.g., verb reinflection), which might not generalize beyond the Arabic language or to more complex linguistic structures.

Moreover, many papers focus on binary gender categories, which excludes non-binary and gender-fluid identities. For example, \citet{stafanovics_mitigating_2020} relies on binary gender annotations that cannot account for non-binary or more complex gender constructs. Similarly, \citet{iluz_exploring_2023}, \citet{costa-jussa_fine-tuning_2020} and \citet{saunders_reducing_2020} focus primarily on binary gender forms, limiting their relevance in scenarios where gender is more fluid. These binary-focused approaches also fail to address how gender bias manifests in languages with non-binary pronouns or where gender is not rigidly categorized, creating an exclusionary gap in research regarding gender diversity. 

Additionally, several theoretical frameworks for addressing bias lack empirical validation, which limits their practical applicability. For example, \citet{kaeser-chen_positionality-aware_2020} offers a framework to account for developers' positionality (i.e., how their background influences the machine learning systems they design) but does not empirically test its effectiveness in real-world MT systems. Similarly, \citet{kruger_outline_2022} and \citet{kruger_outline_2022} introduce an educational model for addressing gender bias in MT but do not present empirical results on the effectiveness of this teaching approach. 

Another recurring challenge is the reliance on small, manually annotated datasets, which restricts the scalability of the findings. \citet{stafanovics_mitigating_2020} relies heavily on manually curated gender annotations, which becomes impractical for larger-scale datasets or for more complex contexts where gender forms may not be binary. Similarly, \citet{levy_collecting_2021} offers a detailed dataset but does not explore how well it scales or generalizes across languages and various social.

\subsection{Outlook for Future Work}
 Future research should aim to generalize findings by investigating how gender bias manifests in broader linguistic structures, such as nouns, verbs, and more complex syntactic forms. Papers like \citet{solmundsdottirmean2022} and \citet{attanasio_tale_2023} could be extended by future research bringing these analyses beyond adjectives and pronouns, respectively. Similarly, studies that focus on specific relational dynamics, like \citet{stewart_whose_2024}, or occupational stereotypes, such as \citet{triboulet_evaluating_2023}, can be used to explore how these biases manifest in other linguistic and social contexts to gain a more comprehensive understanding of MT systems' behavior.

Another key area for future research is the inclusion of non-binary and gender-fluid identities. Studies, such as \citet{stafanovics_mitigating_2020} and \citet{saunders_reducing_2020}, that primarily focus on binary gender distinctions, can be expanded to accommodate non-binary and fluid gender forms that would ensure that bias detection and mitigation methods are more representative of diverse gender identities. Studies on gender coreference and bias resolution, such as \citet{kocmi_gender_2020}, can be adapted to incorporate more inclusive gender categories, ensuring that MT systems are equipped to handle a wider range of gender expressions.

Expanding the scope of research beyond specific languages or language pairs is another crucial area of development. Future research should aim to generalize findings of \citet{prates_assessing_2019} or \citet{costa-jussa_gebiotoolkit_2020} by testing these techniques across more diverse linguistic contexts. This would not only broaden the applicability of bias detection tools but also help ensure that solutions can scale effectively across different language pairs, especially in low-resource and non-Indo-European languages. Similarly, results from language-specific studies like \citet{habash_automatic_2019} and \citet{elaraby_gender_2018} can be used to extend the focus to other languages, allowing researchers to evaluate whether the same gender bias patterns are found across various linguistic systems.

Another priority for future research is improving dataset scalability and addressing the limitations of manual annotations. Papers like \citet{stafanovics_mitigating_2020} and \citet{levy_collecting_2021} would inspire ways to automate the annotation process and build larger, more diverse datasets that better reflect real-world usage. Future research should could expand gender-balanced datasets like \citet{costa-jussa_gebiotoolkit_2020} to cover more diverse text genres, such as news articles, dialogue, and social media, to further improve the robustness of MT systems in handling gender bias.

Improving evaluation metrics and expanding bias analysis beyond specific translation stages is another important area for future work. The methodologies employed in the studies, such as \citet{campolungo_dibimt_2022} and \citet{freitag_ape_2019}, could inspire future research to focus on developing more comprehensive evaluation frameworks that capture gender bias at both the lexical and sentence levels, as well as biases introduced during different stages of the translation process, from input to post-editing. Extending causal models used in papers like \citet{shi_approximating_2023} to address bias directly could also lead to more evaluations of translation quality in relation to gender.

Future work should be expanded to address other types of bias, such as racial or cultural biases. For instance, research could build on the foundational work of studies like \citet{escude_font_equalizing_2019} and \citet{moryossef_filling_2019} by exploring how similar debiasing techniques can be applied across various biases and linguistic contexts. Expanding these techniques would contribute to more comprehensive and effective bias mitigation strategies in machine translation systems.

Testing of theoretical studies in real-world systems such as \citet{kaeser-chen_positionality-aware_2020} that offers a conceptual framework for addressing how developers' positionality influences MT systems, would be beneficial. Similarly, educational frameworks like \citet{kruger_outline_2022} should be evaluated in classroom settings to determine their impact on students' ability to detect and mitigate bias in machine translation. Empirical validation of these theoretical frameworks would provide valuable insights into their practical applications and help shape more effective bias mitigation strategies.

\section{Limitations}
In our approach, we concentrated on African and European written languages, by using a fixed list. Even though we used a two-step approach, we cannot be sure that all relevant languages have been included, as the lists were generated and not derived from a trusted source. For example, Afrikaans was considered as a European language, due to its linguistic origin. However, it could be argued to be added to the list of African languages as this is where it is primarily spoken. At the same time, while focusing on written languages, we are missing out dialects and regional variations of languages. With our focus on European and African languages, we are missing out languages from other regions, and encourage similar work to be applied for those.

\bibliography{custom}

\appendix

\section{Queries and Lists of Languages}
\label{sec:appendix1}

Version: ChatGPT4o

Initial Query: \emph{List of all written African languages?}

Follow-up Query: \emph{Same list in format of python list, only language names.}

Follow-up Query: \emph{Same for European.}

Initial list of languages (African):
Arabic, Amharic, Tigrinya, Hebrew, Tamazight, Kabyle, Shilha, Hausa, Somali, Oromo, Swahili, Zulu, Xhosa, Shona, Kinyarwanda, Kirundi, Lingala, Setswana, Sesotho, Chichewa, Kikuyu, Bemba, Luganda, Bambara, Mandinka, Dyula, Soninke, Akan, Twi, Fante, Ewe, Yoruba, Igbo, Nubian, Dinka, Nuer, Luo, Kanuri, Nama, Khoekhoe, Afrikaans, Cape Verdean Creole, Seychellois Creole, Mauritian Creole, Papiamento, Old Nubian, Ancient Egyptian, Coptic

Initial list of languages (European): 
German, Dutch, Swedish, Danish, Norwegian, Icelandic, Afrikaans, Frisian, Spanish, French, Portuguese, Italian, Romanian, Catalan, Galician, Occitan, Russian, Polish, Czech, Slovak, Ukrainian, Bulgarian, Serbian, Croatian, Bosnian, Slovenian, Belarusian, Macedonian, Montenegrin, Irish, Scottish Gaelic, Welsh, Breton, Cornish, Manx, Greek, Latvian, Lithuanian, Finnish, Estonian, Hungarian, Albanian, Armenian, Basque

Validation Query (new session): \emph{In the two lists above, are there any African or European languages missing?}

Languages added after second query in a new session:

African: Tshiluba, Sango, Malagasy, Wolof, Fon

European: Maltese, Luxembourgish, Sardinian, Faroese, Sorbian, Kashubian, Romansh, Asturian, Aromanian, Aragonese

These combined lists were used to make the selection of the papers collected with Web of Science, with a script. 

This was also the basis for the queries on the ACL Anthology, as shown in the following figures: 

\begin{tcolorbox}[title=ACL Query African Languages]
machine learning \textcolor{green}{AND} bias \textcolor{green}{AND} 
("Arabic" \textcolor{green}{OR} "Amharic" \textcolor{green}{OR} "Tigrinya" \textcolor{green}{OR} "Hebrew" \textcolor{green}{OR} "Tamazight" \textcolor{green}{OR} "Kabyle" \textcolor{green}{OR} "Shilha" \textcolor{green}{OR} "Hausa" \textcolor{green}{OR} "Somali" \textcolor{green}{OR} "Oromo" \textcolor{green}{OR} "Swahili" \textcolor{green}{OR} "Zulu" \textcolor{green}{OR} "Xhosa" \textcolor{green}{OR} "Shona" \textcolor{green}{OR} "Kinyarwanda" \textcolor{green}{OR} "Kirundi" \textcolor{green}{OR} "Lingala" \textcolor{green}{OR} "Setswana" \textcolor{green}{OR} "Sesotho" \textcolor{green}{OR} "Chichewa" \textcolor{green}{OR} "Kikuyu" \textcolor{green}{OR} "Bemba" \textcolor{green}{OR} "Luganda" \textcolor{green}{OR} "Bambara" \textcolor{green}{OR} "Mandinka" \textcolor{green}{OR} "Dyula" \textcolor{green}{OR} "Soninke" \textcolor{green}{OR} "Akan" \textcolor{green}{OR} "Twi" \textcolor{green}{OR} "Fante" \textcolor{green}{OR} "Ewe" \textcolor{green}{OR} "Yoruba" \textcolor{green}{OR} "Igbo" \textcolor{green}{OR} "Nubian" \textcolor{green}{OR} "Dinka" \textcolor{green}{OR} "Nuer" \textcolor{green}{OR} "Luo" \textcolor{green}{OR} "Kanuri" \textcolor{green}{OR} "Nama" \textcolor{green}{OR} "Khoekhoe" \textcolor{green}{OR} "Afrikaans" \textcolor{green}{OR} "Cape Verdean Creole" \textcolor{green}{OR} "Seychellois Creole" \textcolor{green}{OR} "Mauritian Creole" \textcolor{green}{OR} "Papiamento" \textcolor{green}{OR} "Old Nubian" \textcolor{green}{OR} "Ancient Egyptian" \textcolor{green}{OR} "Coptic" \textcolor{green}{OR} “Tshiluba” \textcolor{green}{OR} “Sango” \textcolor{green}{OR} “Malagasy” \textcolor{green}{OR} “Wolof” \textcolor{green}{OR} “Fon”)
\end{tcolorbox}

\captionof{figure}{ACL Query African Languages.}

\begin{tcolorbox}[title=ACL Query European Languages:]
machine learning \textcolor{green}{AND} bias \textcolor{green}{AND} 
("German" \textcolor{green}{OR} "Dutch" \textcolor{green}{OR} "Swedish" \textcolor{green}{OR} "Danish" \textcolor{green}{OR} "Norwegian" \textcolor{green}{OR} "Icelandic" \textcolor{green}{OR} "Afrikaans" \textcolor{green}{OR} "Frisian" \textcolor{green}{OR} "Spanish" \textcolor{green}{OR} "French" \textcolor{green}{OR} "Portuguese" \textcolor{green}{OR} "Italian" \textcolor{green}{OR} "Romanian" \textcolor{green}{OR} "Catalan" \textcolor{green}{OR} "Galician" \textcolor{green}{OR} "Occitan" \textcolor{green}{OR} "Russian" \textcolor{green}{OR} "Polish" \textcolor{green}{OR} "Czech" \textcolor{green}{OR} "Slovak" \textcolor{green}{OR} "Ukrainian" \textcolor{green}{OR} "Bulgarian" \textcolor{green}{OR} "Serbian" \textcolor{green}{OR} "Croatian" \textcolor{green}{OR} "Bosnian" \textcolor{green}{OR} "Slovenian" \textcolor{green}{OR} "Belarusian" \textcolor{green}{OR} "Macedonian" \textcolor{green}{OR} "Montenegrin" \textcolor{green}{OR} "Irish" \textcolor{green}{OR} "Scottish Gaelic" \textcolor{green}{OR} "Welsh" \textcolor{green}{OR} "Breton" \textcolor{green}{OR} "Cornish" \textcolor{green}{OR} "Manx" \textcolor{green}{OR} "Greek" \textcolor{green}{OR} "Latvian" \textcolor{green}{OR} "Lithuanian" \textcolor{green}{OR} "Finnish" \textcolor{green}{OR} "Estonian" \textcolor{green}{OR} "Hungarian" \textcolor{green}{OR} "Albanian" \textcolor{green}{OR} "Armenian" \textcolor{green}{OR} "Basque" \textcolor{green}{OR} “Maltese” \textcolor{green}{OR} “Luxembourgish” \textcolor{green}{OR} “Sardinian” \textcolor{green}{OR} “Faroese” \textcolor{green}{OR} “Sorbian” \textcolor{green}{OR} “Kashubian” \textcolor{green}{OR} “Romansh” \textcolor{green}{OR} “Asturian” \textcolor{green}{OR} “Aromanian” \textcolor{green}{OR} “Aragonese”)
\end{tcolorbox}
\captionof{figure}{ACL Query European Languages.}

\onecolumn

\section{Inclusion and Exclusion Criteria}
\label{sec:appendix2}

The exclusion criteria eliminate papers that are outside the core focus of gender bias in MT. Studies that focus on other areas of NLP, such as speech recognition, grammatical error correction, or sentiment analysis, without addressing MT-specific bias, are excluded. Health-related applications and technical innovations in MT that do not focus on bias are also excluded. 
\begin{longtable}{|p{4cm}|p{8cm}|p{4cm}|}
\hline
\textbf{Author(s)} & \textbf{Paper Title} & \textbf{Reason for Exclusion} \\
\hline
Bircan, Tuba; Ceylan, Duha & \textit{Machine Discriminating: Automated Speech Recognition Biases in Refugee Interviews} & \multirow{3}{4cm}{Focus on ASR systems or speech-based applications} \\
\cline{1-2}
Paulik, M. et al. & \textit{Speech recognition in human mediated translation scenarios} & \\
\cline{1-2}
Casanova, Edresson et al. & \textit{Deep Learning against COVID-19: Respiratory Insufficiency Detection in Brazilian Portuguese Speech} & \\
\hline

Solyman, Aiman et al. & \textit{Automatic Arabic Grammatical Error Correction based on Expectation-Maximization routing and target-bidirectional agreement} & \multirow{2}{4cm}{Focus on grammar correction systems} \\
\cline{1-2}
Mahmoud, Zeinab et al. & \textit{Semi-supervised learning and bidirectional decoding for effective grammar correction in low-resource scenarios} & \\
\hline

Herrera-Espejel, Paula Sofia et al. & \textit{The Use of Machine Translation for Outreach and Health Communication in Epidemiology and Public Health: Scoping Review} & \multirow{2}{4cm}{Focus on public health and health-related applications} \\
\cline{1-2}
Petrova-Antonova, Dessislava et al. & \textit{CogniSoft: A Platform for the Automation of Cognitive Assessment and Rehabilitation of Multiple Sclerosis} & \\
\hline

Wambsganss, Thiemo et al. & \textit{Bias at a Second Glance: A Deep Dive into Bias for German Educational Peer-Review Data Modeling} & \multirow{3}{4cm}{Focus on educational or clinical biases} \\
\cline{1-2}
Pedersen, Jannik et al. & \textit{Investigating anatomical bias in clinical machine learning algorithms} & \\
\cline{1-2}
Rynkiewicz, Agnieszka et al. & \textit{An investigation of the 'female camouflage effect' in autism using a computerized ADOS-2 and a test of sex/gender differences} & \\
\hline

Câmara, António et al. & \textit{Mapping the Multilingual Margins: Intersectional Biases of Sentiment Analysis Systems in English, Spanish, and Arabic} & \multirow{4}{4cm}{Focus on bias in sentiment analysis or NLP tasks beyond machine translation} \\
\cline{1-2}
Chávez Mulsa, Rodrigo Alejandro et al. & \textit{Evaluating Bias In Dutch Word Embeddings} & \\
\cline{1-2}
Adewumi, Tosin et al. & \textit{Bipol: Multi-Axes Evaluation of Bias with Explainability in Benchmark Datasets} & \\
\cline{1-2}
Le, Dieu-thu et al. & \textit{Reducing cohort bias in natural language understanding systems with targeted self-training scheme} & \\
\hline

Zhou, Lei et al. & \textit{Inference Discrepancy Based Curriculum Learning for Neural Machine Translation} & \multirow{7}{4cm}{Focus on specific machine translation advancements or innovations} \\
\cline{1-2}
Seedat, Nabeel et al. & \textit{PEMS: Custom Neural Machine Translation System for Portuguese TV subtitling} & \\
\cline{1-2}
Yu, Heng et al. & \textit{A2R2: Robust Unsupervised Neural Machine Translation With Adversarial Attack and Regularization} & \\
\cline{1-2}
Yan, Jianhao et al. & \textit{Multi-Unit Transformers for Neural Machine Translation} & \\
\cline{1-2}
Zhang, Tianfu et al. & \textit{Enlivening Redundant Heads in Multi-head Self-attention for Machine Translation} & \\
\cline{1-2}
Liu, Yijin et al. & \textit{Confidence-Aware Scheduled Sampling for Neural Machine Translation} & \\
\cline{1-2}
Stahlberg, Felix et al. & \textit{On NMT Search Errors and Model Errors: Cat Got Your Tongue?} & \\
\hline

Amrhein, Chantal et al. & \textit{Exploiting Biased Models to De-bias Text: A Gender-Fair Rewriting Model} & \multirow{3}{4cm}{Focus on debiasing or bias mitigation in NLP} \\
\cline{1-2}
Goodman, Sebastian et al. & \textit{TeaForN: Teacher-Forcing with N-grams} & \\
\cline{1-2}
Helal, Mohammed et al. & \textit{The CyberEquity Lab at FIGNEWS 2024 Shared Task: Annotating a Corpus of Facebook Posts to Label Bias and Propaganda in Gaza-Israel War Coverage in Five Languages} & \\
\hline

\end{longtable}

The inclusion criteria for the manual selection process focus on papers that include at least one language from the query and specifically address gender bias in machine translation (MT). This includes studies that explore bias detection and mitigation techniques in MT systems, particularly in multilingual settings or languages with grammatical gender. Papers proposing novel approaches to reducing bias, such as gender-aware learning or the use of large-scale datasets for bias evaluation, are prioritized. We incorporated a theoretical paper that addresses the issue of positionality in machine translation despite it not covering any of the languages in our query.
\begin{longtable}{|p{4cm}|p{8cm}|p{4cm}|}
\hline
\textbf{Author(s)} & \textbf{Paper Title} & \textbf{Reason for Inclusion} \\
\hline

\citet{solmundsdottirmean2022} & \textit{Mean Machine Translations: On Gender Bias in Icelandic Machine Translations} & \multirow{5}{4cm}{\textbf{General Approaches \& Evaluations of Gender Bias in MT}} \\
\cline{1-2}
\citet{savoldi_gender_2021} & \textit{Gender Bias in Machine Translation} & \\
\cline{1-2}
\citet{costa-jussa_fine-tuning_2020} & \textit{Fine-tuning Neural Machine Translation on Gender-Balanced Datasets} & \\
\cline{1-2}
\citet{vashishtha_evaluating_2023} & \textit{On Evaluating and Mitigating Gender Biases in Multilingual Settings} & \\
\cline{1-2}
\citet{leetargetagnostic2023} & \textit{Target-Agnostic Gender-Aware Contrastive Learning for Mitigating Bias in Multilingual Machine Translation} & \\
\hline

\citet{vanmassenhove_machine_2021} & \textit{Machine Translationese: Effects of Algorithmic Bias on Linguistic Complexity in Machine Translation} & \multirow{3}{4cm}{\textbf{Bias Amplification \& Linguistic Impact in MT}} \\
\cline{1-2}
\citet{kocmi_gender_2020} & \textit{Gender Coreference and Bias Evaluation at WMT 2020} & \\
\cline{1-2}
\citet{cabrera_gender_2023} & \textit{Gender Lost In Translation: How Bridging The Gap Between Languages Affects Gender Bias in Zero-Shot Multilingual Translation} & \\
\hline

\citet{stewart_whose_2024} & \textit{Whose wife is it anyway? Assessing bias against same-gender relationships in machine translation} & \multirow{2}{4cm}{\textbf{Bias Against Same-Gender Relationships}} \\
\cline{1-2}
\citet{attanasio_tale_2023} & \textit{A Tale of Pronouns: Interpretability Informs Gender Bias Mitigation for Fairer Instruction-Tuned Machine Translation} & \\
\hline

\citet{mash_unmasking_2024} & \textit{Unmasking Biases: Exploring Gender Bias in English-Catalan Machine Translation through Tokenization Analysis and Novel Dataset} & \multirow{2}{4cm}{\textbf{Tokenization \& Subword Units Impact on Gender Bias}} \\
\cline{1-2}
\citet{iluz_exploring_2023} & \textit{Exploring the Impact of Training Data Distribution and Subword Tokenization on Gender Bias in Machine Translation} & \\
\hline

\citet{levy_collecting_2021} & \textit{Collecting a Large-Scale Gender Bias Dataset for Coreference Resolution and Machine Translation} & \multirow{2}{4cm}{\textbf{Large-Scale Datasets for Evaluating Gender Bias in MT}} \\
\cline{1-2}
\citet{habash_automatic_2019} & \textit{Automatic Gender Identification and Reinflection in Arabic} & \\
\hline

\end{longtable}

\twocolumn

\end{document}